\documentclass[journal,comsoc]{IEEEtran}
\usepackage[T1]{fontenc}
\usepackage{cite}

\ifCLASSINFOpdf
   \usepackage[pdftex]{graphicx}
\else
   \usepackage[dvips]{graphicx}
\fi

\usepackage{amsmath}
\usepackage[cmintegrals]{newtxmath}

\usepackage[numbers]{natbib}
\usepackage{booktabs}
\usepackage{multirow}
\usepackage{multicol}
\usepackage{amsmath}
\usepackage{algpseudocode}
\usepackage{mathtools}
\usepackage[linesnumbered,ruled,vlined]{algorithm2e}

\mathchardef\mhyphen="2D

\begin{document}
%
\title{Actively Obtaining Environmental Feedback for Autonomous Action Evaluation Without Predefined Measurements}

\author{Hong~Su
\IEEEcompsocitemizethanks{\IEEEcompsocthanksitem H. Su is with the School of Computer Science, Chengdu University of Information Technology, Chengdu, China.\\
 E-mail: suguest@126.com. \\
\protect\\
}
\thanks{}}

\markboth{Journal of \LaTeX\ Class Files,~Vol.~14, No.~8, August~2015}%
{Shell \MakeLowercase{\textit{et al.}}: Bare Demo of IEEEtran.cls for IEEE Communications Society Journals}
%

\maketitle

\begin{abstract}

Obtaining reliable feedback from the environment is a fundamental capability for intelligent agents to evaluate the correctness of their actions and to accumulate reusable knowledge. However, most existing approaches rely on predefined measurements or fixed reward signals, which limits their applicability in open-ended and dynamic environments where new actions may require previously unknown forms of feedback. 
To address these limitations, this paper proposes an Actively Feedback Getting model, in which an AI agent proactively interacts with the environment to discover, screen, and verify feedback without relying on predefined measurements. Rather than assuming explicit feedback definitions, the proposed method exploits action-induced environmental differences to identify target feedback that is not specified in advance, based on the observation that actions inevitably produce measurable changes in the environment. In addition, a self-triggering mechanism, driven by internal objectives such as improved accuracy, precision, and efficiency, is introduced to autonomously plan and adjust actions, thereby enabling faster and more focused feedback acquisition without external commands.
Experimental results demonstrate that the proposed active approach significantly improves the efficiency and robustness of factor identification. 

\end{abstract}


\IEEEpeerreviewmaketitle

\section{Introduction}
\label{sec_introduction}

Large language models (LLMs) have been widely adopted as core components of AI agents operating in real-world environments \cite{huang2025recommender} \cite{gao2024large}.
By providing strong reasoning, planning, and language understanding capabilities, LLMs enable agents to interpret complex situations and generate sophisticated action plans.
As a result, LLM-powered agents are increasingly deployed in interactive systems, autonomous control, and decision-support applications.

Despite these advances, most existing AI agents in real-world settings still rely heavily on \emph{predefined measurements or actions}, explicit reward functions, or externally specified feedback signals to evaluate the outcomes of their actions.
Such designs impose strong assumptions on how feedback should be represented and interpreted, which significantly limits agent autonomy in open-ended and dynamic environments.
When new actions are introduced or when relevant feedback does not conform to predefined forms, agents often lack effective mechanisms to discover, interpret, and verify feedback.

In contrast, human behavior in natural environments does not depend on predefined measurement schemas.
Humans actively interact with the environment, observe differences caused by their actions, and deliberately perform additional actions to identify, amplify, or accelerate feedback.
Rather than passively waiting for results, humans intervene in the feedback generation process, adjusting actions and observation scopes to obtain clearer and more reliable feedback.
This ability to actively obtain feedback is a key factor underlying human adaptability and learning.

Inspired by this observation, we argue that AI agents should not rely solely on passive LLM-driven reasoning or externally defined feedback channels.
Instead, agents should be equipped with mechanisms to actively discover feedback from the environment, even when no explicit measurement is provided in advance.
In particular, action-induced \emph{environmental differences} provide a universal signal through which feedback can be detected, screened, and validated.
Moreover, agents should be able to plan and execute actions that intentionally expose or accelerate feedback, guided by internal objectives (as in Section \ref{sec_int_tri}) rather than external commands.

In this paper, we propose an \emph{Actively Feedback Getting} model that enables AI agents to autonomously obtain, validate, and accumulate feedback from the environment without relying on predefined measurements.
The proposed model treats difference as the fundamental carrier of feedback, integrates active action as a form of feedback intervention, and employs internal action trigger sources to drive exploration and refinement.
By combining difference-driven detection, active screening and validation, and accumulated learning of action--feedback relationships, the model supports efficient and adaptive feedback acquisition in open-ended environments.

The main contributions of this paper are summarized as follows:
\begin{itemize}
    \item We propose a difference-driven feedback model that enables implicit feedback discovery without assuming predefined measurements.
    \item We introduce active action as a feedback intervention mechanism, allowing agents to accelerate and disambiguate feedback acquisition through deliberate environmental manipulation.
    \item We design an accumulated learning mechanism that records and reuses action--feedback relationships, supporting continual adaptation and efficiency improvement. 
\end{itemize}

The remainder of this paper is organized as follows.
Section \ref{sec_related_works} reviews related work on feedback modeling, active perception, and autonomous agents.
Section \ref{sec_afg_model} presents the proposed Actively Feedback Getting model in detail.
Section \ref{sec_verification} provides experimental verification and analysis.
Finally, Section \ref{sec_conclusion} concludes the paper and discusses future research directions.

\section{Related Works}
\label{sec_related_works}

This section reviews prior work related to feedback modeling, active perception, causal reasoning, and autonomous agents.
We highlight the limitations of existing approaches in open-ended environments and clarify how the proposed model differs from and extends previous studies.

\subsection{Predefined Feedback and Reward Modeling}

Most existing intelligent systems rely on predefined feedback representations, such as explicit reward functions, handcrafted metrics, or task-specific evaluation signals.
Reinforcement learning (RL) frameworks typically assume that a scalar reward is available after each action or episode, enabling policy optimization through repeated interaction~\cite{shakya2023reinforcement} \cite{matsuo2022deep}.
Similarly, many decision-making and control systems depend on predefined measurements to assess action outcomes.

While effective in closed or well-defined environments, such approaches face challenges in open-ended real-world scenarios.
Designing appropriate reward functions or measurement criteria requires strong prior knowledge and often fails to generalize when new actions, goals, or environmental dynamics emerge.
The proposed model differs by removing the assumption of predefined measurements and treating feedback as an implicit signal discovered through interaction.

\subsection{Active Perception and Active Learning}

Active perception emphasizes that an agent can improve its understanding of the environment by actively selecting actions that influence observations~\cite{bajcsy2018revisiting}.
Similarly, active learning focuses on querying informative data points to reduce uncertainty in supervised or semi-supervised settings.

These approaches demonstrate that action selection can significantly affect information acquisition.
However, most active perception and learning methods still assume predefined observation features, labels, or uncertainty measures.
In contrast, our model extends the notion of activeness to the feedback level, enabling agents to actively intervene in the environment to discover and validate feedback itself, even when its form is unknown in advance.

\subsection{Causal Discovery and Difference-Based Reasoning}

Causal discovery methods aim to identify cause--effect relationships from observational or interventional data~\cite{nogueira2022methods} \cite{zheng2024causal}.
Difference-based reasoning, including counterfactual analysis, has been used to isolate influential factors by comparing alternative conditions or outcomes.

Although these methods provide valuable theoretical tools, they often assume access to explicit variables, structured datasets, or well-defined interventions.
Moreover, many causal approaches focus on offline analysis rather than online interaction.
The proposed model adopts the intuition of difference-based reasoning but integrates it into an online, action-driven feedback acquisition process without requiring explicit causal variable definitions.

\subsection{Autonomous Agents and Intrinsic Motivation}

Recent studies on autonomous agents emphasize intrinsic motivation, self-goal-driven behavior, and internal action triggers~\cite{wang2024survey}.
Such mechanisms allow agents to explore environments beyond externally specified objectives and improve long-term adaptability.

Large language models have further enhanced agent autonomy by enabling high-level reasoning, planning, and instruction following.
However, LLM-driven agents often remain passive with respect to feedback acquisition, relying on externally provided observations or predefined evaluation criteria.
Our work complements these efforts by introducing internal action trigger sources specifically designed for feedback discovery and validation, enabling agents to actively shape the conditions under which feedback emerges.

\subsection{Summary}

In summary, prior work has explored predefined feedback modeling, active information acquisition, causal reasoning, and autonomous behavior from different perspectives.
However, existing approaches typically assume explicit feedback representations or passive observation mechanisms.
The proposed Actively Feedback Getting model differs by unifying difference-driven feedback detection, active action intervention, and accumulated learning into a coherent framework that supports feedback acquisition without predefined measurements in open-ended environments.

\section{Actively Feedback Getting Model}
\label{sec_afg_model}

\begin{figure}
    \includegraphics[width=3.5in]{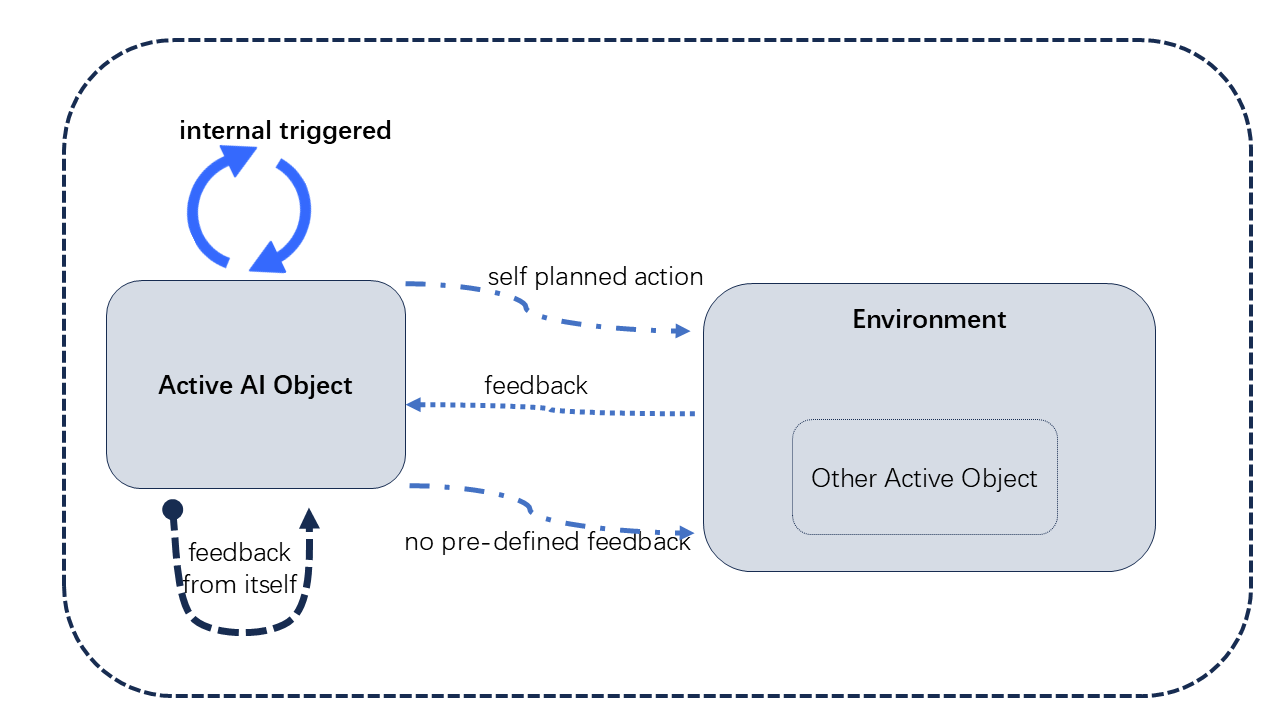}
    \caption{Overview of the proposed Actively Feedback Getting model}
    \label{fig_archetecture}
\end{figure}

The Actively Feedback Getting model aims to enable an AI system to interact with the environment without relying on predefined measurements or externally specified feedback signals.
Instead of passively receiving evaluation criteria, the agent autonomously initiates actions and interactions with the environment, driven by internal objectives.
Through such internally triggered interactions, the agent actively discovers, validates, and accumulates feedback, as illustrated in Fig.~\ref{fig_archetecture}.

\subsection{Problem Definition and Assumptions}
\label{sec_problem_definition}

We consider an intelligent agent interacting with an environment through actions and observations. 
Let the environment state be denoted as $E_t$ at time $t$, and let the agent perform an action $a_t$, which may induce a state transition to $E_{t+1}$. 
Unlike traditional reinforcement learning or feedback-driven systems, we do not assume the existence of predefined measurements, reward functions, or explicit feedback signals.

The objective of the agent is to evaluate the correctness and impact of its actions and to accumulate reusable knowledge from the environment, even when the form of feedback is unknown in advance.
Formally, the agent aims to identify a feedback signal $f_t$ associated with action $a_t$, where $f_t$ is not specified a priori and may emerge implicitly from environmental changes.

We base our model on the following assumptions:

\begin{itemize}
    \item \textbf{Action-induced change}: Any action performed by the agent inevitably causes a change in the environment, either immediately or after a temporal delay.
    \item \textbf{Observable difference}: Environmental changes caused by actions can be observed as differences between states across time or space, even if their semantic meaning is not explicitly defined.
    \item \textbf{No predefined measurement}: The agent does not rely on predefined metrics, reward signals, or labeled feedback to assess action outcomes.
    \item \textbf{Internal autonomy}: The agent is capable of autonomously deciding whether to initiate actions based on internal objectives, rather than external commands.
\end{itemize}

Under these assumptions, the core challenge is to determine how an agent can \emph{actively obtain} meaningful feedback from the environment, distinguish relevant feedback from irrelevant variations, and verify its correctness without relying on predefined measurement criteria.
The proposed Actively Feedback Getting model addresses this challenge by treating \emph{difference} as a universal carrier of feedback and by enabling the agent to proactively plan actions that expose such differences.

\subsection{Difference-Driven Feedback Detection}
\label{sec_difference_detection}

In the absence of predefined measurements, the proposed model treats \emph{difference} as the primary carrier of feedback.
The underlying observation is that any action performed by an agent inevitably induces changes in the environment, and such changes can be identified through differences between environmental states.

Formally, let $E_t$ and $E_{t+\Delta}$ denote the environment states before and after an action (or a sequence of actions) within a given temporal or spatial scope.
The environmental difference is defined as
\begin{equation}
    \Delta E = \mathcal{D}(E_t, E_{t+\Delta}),
\end{equation}
where $\mathcal{D}(\cdot)$ denotes a generic difference operator that does not rely on predefined measurement forms.
This operator may capture variations in observable variables, structural configurations, temporal patterns, or other detectable environmental changes.

Unlike traditional approaches that depend on explicit metrics or reward signals, the difference-driven mechanism enables implicit feedback identification.
If no difference is detected within the current observation scope, the agent may infer that the action has no observable impact under existing conditions, or that the feedback is delayed or masked.
Conversely, detected differences indicate potential feedback candidates associated with the executed action.

Differences may manifest along multiple dimensions, including:
\begin{itemize}
    \item \textbf{Temporal differences}, where changes emerge after a time delay following an action;
    \item \textbf{Spatial differences}, where variations occur across different regions or objects;
    \item \textbf{Magnitude differences}, reflecting the intensity or scale of change;
    \item \textbf{Frequency differences}, indicating repeated or persistent changes over time.
\end{itemize}

\subsubsection{Degree of Difference}
\label{sec_degree_difference}

Not all differences carry the same informational value.
To characterize feedback strength, we introduce the notion of \emph{degree of difference}, which reflects how significant a detected difference is.
The degree may be quantified by multiple factors, such as duration, intensity, frequency of occurrence, or deviation from normal behavior.

Let $\delta \in \Delta E$ denote an individual detected difference.
Its degree can be represented by a scalar score
\begin{equation}
    s(\delta) = g(\delta),
\end{equation}
where $g(\cdot)$ is an evaluation function derived from sensory measurements, historical statistics, or large language model (LLM) reasoning.
In practice, detecting and estimating $s(\delta)$ may require the agent to perform additional actions, deploy sensors, or conduct controlled measurements.

Frequency also contributes to the degree of difference.
Differences that occur repeatedly or persist over time are generally considered more informative than sporadic ones.
LLMs may further assist in assessing difference degree by incorporating contextual knowledge and prior experience.

Based on the degree of difference, feedback can be categorized as minor, significant, or abnormal.
Abnormal differences correspond to deviations that exceed a predefined threshold or fall outside previously observed patterns, indicating potentially novel or critical feedback.

\subsubsection{Initial Screening Mechanism}
\label{sec_initial_screening}

It is important to note that not all detected differences correspond to meaningful feedback.
Some differences may arise from irrelevant variations, environmental noise, or disturbing factors.
Therefore, difference-driven detection serves as an \emph{initial screening mechanism}, whose output requires further refinement.

The candidate differences identified at this stage are subsequently processed through active feedback screening, action adjustment, and logical evaluation, as described in the following subsections.
By elevating difference to a fundamental abstraction while deferring semantic interpretation, the proposed model enables flexible and measurement-free feedback acquisition in open-ended and dynamic environments.

\subsection{Active Feedback Screening and Validation}
The primary objective of feedback acquisition is to obtain \emph{correct and reliable} feedback from the environment.
This objective functions as an internal action trigger source (Section~\ref{sec_int_tri}), removing reliance on external instructions and activating an explicit \emph{feedback screening and validation} stage.
Once triggered, the agent evaluates whether observed differences genuinely reflect the effects of its actions rather than noise or irrelevant variations, using either previously learned validation strategies or on-demand reasoning assisted by a large language model (LLM).

\subsubsection{Active Feedback Screening} \label{sec_active_screening}

Difference-driven detection provides a set of candidate feedback signals, but these candidates may include irrelevant or ambiguous variations.
To refine the detected differences and identify feedback that is truly related to the agent's actions, we introduce an \emph{active feedback screening} mechanism.

Active feedback screening refers to the process by which the agent proactively determines \emph{which differences should be regarded as meaningful feedback}.
Instead of passively accepting all observed changes, the agent forms hypotheses about possible feedback and selectively focuses on collecting evidence that supports or refutes these hypotheses.

Two complementary screening mechanisms are employed in the proposed model.

\paragraph{Action--Result Screening}
The first mechanism is based on the direct relationship between actions and environmental changes.
When an action $a_t$ is executed, the agent monitors the environment for subsequent differences and evaluates whether these differences can be causally attributed to $a_t$.
Differences that consistently co-occur with the execution of the same or similar actions are retained as feedback candidates, while those that appear independently of the action are gradually discarded.

\paragraph{Expectation-Guided Screening}
The second mechanism leverages the agent's accumulated experience and large language model (LLM) reasoning capabilities.
Before or during action execution, the agent queries its internal knowledge or an LLM to generate expectations about possible outcomes of the action.
These expectations act as soft templates that guide the screening process, enabling the agent to focus on differences that align with plausible action outcomes.

Expectation-guided screening does not impose predefined measurements.
Instead, it narrows the feedback space by prioritizing differences that are semantically or causally consistent with the predicted effects of the action.

Through active feedback screening, the agent reduces noise in the feedback acquisition process and improves the reliability of subsequent analysis.
This mechanism serves as a bridge between raw difference detection and higher-level action planning and evaluation, which are discussed in the following subsections.

\subsubsection{Correctness Judgment and Logical Evaluation}
\label{sec_correctness_judgment}

Active feedback acquisition does not guarantee that all obtained feedback is correct or meaningful.
Environmental noise, delayed effects, hidden factors, or improper actions may lead to false or misleading feedback.
Therefore, the proposed model incorporates a \emph{correctness judgment and logical evaluation} process at the logical layer to validate both the acquired feedback and the actions that produced it.

\paragraph{Feedback Correctness Judgment}
Once candidate feedback is obtained, the agent evaluates its correctness by examining its consistency with prior observations, accumulated knowledge, and contextual constraints.
Feedback that falls within the expected range of previously observed patterns is considered reliable, while abnormal or novel feedback is treated with caution.
For unfamiliar feedback, the agent may request additional reasoning support from a large language model (LLM) to analyze possible causes, explanations, or confounding factors.

If the feedback cannot be reliably verified, the agent may repeat actions, perform comparative tests, or modify environmental conditions to reduce randomness and ambiguity.
In cases where expected feedback is absent, the agent also analyzes potential reasons for the failure, such as insufficient action strength, limited observation scope, or interference from other factors, and then adjusts actions accordingly.

\paragraph{Action Evaluation and Refinement}
In addition to judging feedback correctness, the agent evaluates the effectiveness of its own actions.
This includes assessing whether the executed actions were sufficient, complete, or properly targeted to expose the intended feedback.
Actions are reviewed step by step to identify possible errors or inefficiencies, and improvements are incorporated into future action planning.

This reflective evaluation enables the agent to refine its behavior over time, gradually reducing unnecessary actions and improving feedback acquisition efficiency.
By jointly evaluating feedback validity and action quality, the logical layer closes the feedback loop between perception and action, ensuring that subsequent learning and decision-making are based on reliable information.

\subsection{Active Action as Feedback Intervention}
\label{sec_active_action_new}

In the proposed model, \emph{active action} is not merely the execution of behaviors to observe outcomes, but a deliberate intervention in the feedback generation process driven by internal objectives.
Rather than passively waiting for feedback to emerge, the agent actively seeks to improve the quality, clarity, and efficiency of feedback acquisition, which serves as an internal action trigger source (Section~\ref{sec_int_tri}).

The primary role of active action is to expose, amplify, or disambiguate feedback when passive observation is insufficient.
Unlike difference detection or feedback screening, which focus on identifying candidate feedback, active action directly manipulates environmental conditions or action parameters to make feedback observable within a controllable scope.

\subsubsection{Active Adjustment for Factor Identification}
\label{sec_active_adjustment}

When feedback is weak, mixed, or hidden by other measurements, the agent performs \emph{active adjustment} to isolate influential factors.
This is achieved by intentionally strengthening, weakening, enabling, or disabling selected action factors and comparing the resulting feedback differences.

Formally, let $a(\lambda)$ denote an action parameterized by a controllable factor $\lambda$.
The agent evaluates the effect of $\lambda$ by comparing feedback outcomes
\[
f(\lambda_1) \quad \text{and} \quad f(\lambda_2),
\]
where $\lambda_1 \neq \lambda_2$.
Consistent changes in feedback across adjustments indicate a causal association between the factor and the observed feedback.
This process allows the agent to identify dominant factors even when direct observation fails.

\subsubsection{Scope Operations for Feedback Discovery}
\label{sec_scope_operation}

In real-world environments, feedback may not be observable within a fixed temporal or spatial range.
Therefore, the agent performs \emph{scope operations} to regulate the observation boundary of feedback acquisition \cite{su2025intuition}.

Scope expansion is applied when no relevant feedback is detected, including extending the temporal window, enlarging the spatial region, or incorporating interactions with additional agents.
Conversely, scope reduction is used to suppress irrelevant variations by focusing on a narrower observation range, enabling finer-grained feedback analysis.

Through dynamic scope adjustment, the agent balances completeness and precision, ensuring that feedback acquisition remains both tractable and informative.

\subsubsection{Action-Level Evaluation and Refinement}
\label{sec_action_evaluation}

Active action also requires evaluation at the action level.
After feedback is obtained, the agent assesses whether the executed actions were appropriate, sufficient, and correctly targeted.
Let $a$ denote the executed action plan (including factor adjustments and scope settings), and let $f$ denote the resulting validated feedback.
We evaluate an action plan by a utility objective
\begin{equation}
    J(a) = \alpha \cdot \mathrm{Rel}(f) \;-\; \beta \cdot \mathrm{Cost}(a) \;-\; \gamma \cdot \mathrm{Amb}(f),
\end{equation}
where $\mathrm{Rel}(f)$ measures feedback reliability/consistency, $\mathrm{Cost}(a)$ measures action cost (e.g., time, number of interventions), and $\mathrm{Amb}(f)$ measures ambiguity (e.g., sensitivity to disturbing factors). $\alpha,\beta,\gamma>0$ are weights.

Actions are reviewed retrospectively by comparing the observed outcome with plausible alternatives.
Given a set of candidate alternative plans $\mathcal{A}$ (e.g., different adjustment strengths or observation scopes), the agent selects a refined plan
\begin{equation}
    a^{*} = \arg\max_{a' \in \mathcal{A}} \mathbb{E}\!\left[J(a')\right],
\end{equation}
where the expectation is taken over environmental randomness and unobserved confounders.

If ambiguity remains high (i.e., $\mathrm{Amb}(f)$ exceeds a threshold), the agent repeats the intervention under modified conditions or consults an LLM to propose additional discriminative adjustments that reduce confounding effects.
This evaluation-and-update loop enables continual refinement of intervention strategies, ensuring that future active actions become progressively more efficient and reliable.

Overall, active action transforms feedback acquisition from passive observation into a controlled experimental process, allowing the agent to actively shape the conditions under which meaningful feedback emerges.

\subsection{Accumulated Learning of Action--Feedback Relationships}
\label{sec_accumulated_learning}

To enable long-term improvement and reuse of acquired knowledge, the proposed model incorporates an accumulated learning mechanism that records and organizes relationships between actions and their corresponding feedback.
These relationships serve as a reference for future feedback screening, action planning, and correctness judgment.

\subsubsection{Action--Feedback Relationship Representation}
An action--feedback relationship is represented as a mapping between an action (or a set of action-related factors) and the observed feedback:
\begin{equation}
    a \;\rightarrow\; f,
\end{equation}
where $a$ denotes the action context and $f$ denotes the validated feedback.
Unlike traditional learning methods that require large amounts of data, this representation allows the agent to retain relationships derived from sparse or rare events.

\subsubsection{Obvious Recording and Mixed Memory}
\label{sec_obvious_mixed_memory}

For frequently occurring and regular interaction patterns, the agent relies on deep learning mechanisms to form generalized representations of normal environmental behavior.
Let $\mathcal{D} = \{(a_i, f_i)\}$ denote the set of observed action--feedback pairs.
When the empirical occurrence probability $P(a,f)$ is sufficiently high, these pairs are absorbed into a parametric model $M_{\theta}$ through statistical learning, enabling generalization across similar states and actions.

However, for rare, abnormal, or previously unseen situations—where $P(a,f)$ is low or undefined—statistical learning alone becomes ineffective or inefficient.
In such cases, the agent adopts an \emph{obvious recording} strategy~\cite{su2025human}, explicitly storing salient action--feedback relationships in a symbolic form:
\begin{equation}
    \mathcal{R}_{\text{obs}} = \{ a \rightarrow f \mid P(a,f) < \epsilon \},
\end{equation}
where $\epsilon$ is a small threshold indicating insufficient statistical support.

The resulting memory system is a \emph{mixed memory architecture}, consisting of a parametric memory $M_{\theta}$ for common patterns and a non-parametric memory $\mathcal{R}_{\text{obs}}$ for sparse or exceptional cases.
This hybrid design enables the agent to achieve both robustness—through statistical generalization—and adaptability—through precise retention of rare but informative experiences.

\subsubsection{Movable Relationships}
Not all recorded relationships are equally reusable across different scenarios.
To avoid overfitting to specific contexts, the model introduces the concept of \emph{movable relationships}, which represent relationships that can be transferred across time, space, or environmental conditions.
Before recording a relationship, the agent evaluates—possibly with the assistance of an LLM—whether the relationship is likely to remain valid in other scenarios.
Only relationships with sufficient generality are promoted to movable relationships.

Movable relationships may exist at different levels of abstraction.
General relationships capture broadly applicable patterns, while more specific relationships include additional contextual attributes.
Together, these relationships form a hierarchical or graph-structured memory, enabling the agent to progressively refine its understanding as new feedback is acquired.

By accumulating and organizing action--feedback relationships in this manner, the proposed model enables continual learning and supports increasingly efficient and reliable feedback acquisition in complex and evolving environments.

\subsubsection{Difference-Centered Memory with Scenario Mapping}
\label{sec_difference_centered_memory}

When accumulating action--feedback relationships, storing all observed information is neither efficient nor necessary.
Many environmental factors remain unchanged across different scenarios and thus contribute little to distinguishing feedback causes.
To address this, we propose a \emph{difference-centered memory} mechanism, in which the most significant difference is used as a key to index and retrieve action--feedback relationships.

Let $\Delta E = \mathcal{D}(E_t, E_{t+\Delta})$ denote the set of detected environmental differences after executing an action.
Each difference $\delta_k \in \Delta E$ is associated with a significance score $s(\delta_k)$, which may reflect magnitude, frequency, persistence, or abnormality.
The agent selects the most informative difference as
\begin{equation}
    \delta^{*} = \arg\max_{\delta_k \in \Delta E} s(\delta_k).
\end{equation}

The selected $\delta^{*}$ serves as a compact and discriminative \emph{memory key}, under the assumption that learning proceeds from identifying one dominant causal factor at a time.
Using this key, the agent stores a mapping to the corresponding action--feedback relationship and its contextual scenario:
\begin{equation}
    \delta^{*} \;\rightarrow\; \langle a, f, \mathcal{S} \rangle,
\end{equation}
where $a$ denotes the action (or action factors), $f$ denotes the validated feedback, and $\mathcal{S}$ represents the scenario context, including temporal, spatial, or environmental attributes.

By centering memory on the most distinctive difference, the agent avoids redundant storage of common or irrelevant factors.
When encountering a new scenario, the agent can retrieve past experiences by matching observed differences to stored keys, enabling efficient feedback prediction and action guidance.
This mechanism also supports incremental refinement: if the retrieved relationship proves insufficient in a new context, additional distinguishing differences can be incorporated to form more specific keys, gradually enriching the memory structure.

\subsection{Source of Active Feedback}
\label{sec_feedback_source}

In the proposed model, feedback may originate from multiple sources rather than a single predefined channel.
Specifically, feedback can be obtained from the environment, from other intelligent agents, and from the agent itself.
These heterogeneous sources jointly support active feedback acquisition and internal action triggering.

For non-intelligent environments, the relationship between actions and feedback is governed primarily by physical rules.
In such cases, feedback emerges directly from action-induced environmental changes without semantic interpretation or intentional response.
The difference-driven mechanism discussed earlier enables the agent to extract feedback from these changes.

In contrast, intelligent agents may actively respond to actions, thereby altering the feedback itself.
For example, when an agent interferes with another intelligent agent, the latter may change its behavior instead of remaining passive.
Such interactions can amplify or reshape feedback and must be considered as part of the feedback acquisition process.

\subsubsection{Feedback from Other Intelligent Agents}

Feedback obtained from other intelligent agents is often indirect and context-dependent.
An observed response may be influenced by additional internal states, goals, or external stimuli.
For instance, a painful stimulus may result in a facial expression of sadness rather than a direct physical signal.
To interpret such feedback correctly, the observing agent may need to expand its observation scope and consider multiple interacting factors.

\subsubsection{Self-Impact Feedback}

An agent may also receive feedback through the impact of its actions on itself.
This self-impact feedback requires internal sensing mechanisms to detect changes in internal states, such as discomfort, resource consumption, or performance degradation.
The degree of self-impact may be quantified through intensity, duration, or frequency of occurrence.
For example, stronger or more persistent negative effects typically indicate more significant feedback.

\subsubsection{Internal Action Trigger Source} \label{sec_int_tri}

Actions in the proposed model are not executed solely in response to external commands.
Instead, they are initiated by \emph{internal action trigger sources}, which determine whether and when further actions should be performed.
Internal triggers may arise from the agent's goals (e.g., improving performance or precision), from environmental feedback, from interactions with other agents, or from negative self-impact.

Only when an internal trigger is activated does the agent engage in further active feedback acquisition.
An agent operating under internal action trigger control, rather than predefined action scripts, is considered \emph{active}.
Self-goal-driven behavior has been explored in prior work~\cite{su2025active}, and in this model, feedback itself—especially abnormal or unfavorable feedback—can serve as a trigger for further exploration and adaptation.

\section{Verification}
\label{sec_verification}

In this section, we experimentally verify the effectiveness of the proposed actively feedback getting model.
The evaluation focuses on two core capabilities: 
(1) discovering feedback without predefined measurements, and 
(2) accelerating causal factor identification through active actions.
All experiments are designed to isolate the contribution of difference-driven detection and active intervention mechanisms.

\subsection{Unknown Measurement Discovery via Difference Comparison}
\label{sec_unknown_measurement}

This experiment evaluates whether the proposed difference-driven feedback mechanism enables an agent to identify relevant causal factors when no explicit measurements or predefined criteria are provided.

\subsubsection{Experimental Setup}
We construct a text-based scenario in which multiple environmental and personal factors are implicitly embedded, while only one factor is strongly correlated with the target outcome.
The agent is required to identify the possible reason for the observed outcome solely through content analysis, without access to predefined measurement dimensions or causal hints.

Two reasoning strategies are compared:
\begin{itemize}
    \item $\mathcal{M}_{\text{difference}}$: a difference-oriented strategy that explicitly prompts the agent to compare the target subject with others and focus on distinguishing factors;
    \item $\mathcal{M}_{\text{direct}}$: a direct reasoning strategy that asks for possible reasons without invoking comparison.
\end{itemize}

Both strategies operate on the same input content and are evaluated against a reference explanation representing the most plausible causal factor.

\subsubsection{Prompts and Evaluation Metric}
To ensure a controlled comparison, both strategies use fixed prompt templates.
For $\mathcal{M}_{\text{difference}}$ (the proposed method), an additional comparison instruction is provided, explicitly requiring the agent to identify distinguishing attributes relative to other subjects and to ignore common characteristics.
In contrast, $\mathcal{M}_{\text{direct}}$ receives only a general request to explain the observed outcome.

Formally, the comparison instruction is defined as:
\begin{quote}
\small
\emph{``Compare differences from other children and focus on aspects that differ from most peers. Similarities with other classmates should not be included when identifying the reasons.''}
\end{quote}

The corresponding query templates are:
\begin{itemize}
    \item $Q_{\text{difference}}$: ``Identify the reasons for Xiao Ming's strong innovative abilities based on the following content, using the above comparison instruction.''
    \item $Q_{\text{direct}}$: ``Identify the reasons for Xiao Ming's strong innovative abilities based on the following content.''
\end{itemize}

To quantify effectiveness, we measure the semantic similarity between the agent’s output and a reference explanation $r$, which represents the most plausible causal factor.
Let $o_i$ denote the output generated in the $i$-th trial.
The similarity score is computed as
\begin{equation}
    s_i = \mathrm{Sim}_{\text{MiniLM}}(o_i, r),
\end{equation}
where $\mathrm{Sim}_{\text{MiniLM}}(\cdot,\cdot)$ denotes a sentence-level semantic similarity function implemented using the pretrained language model \textit{all-MiniLM-L6-v2}.
Higher similarity scores indicate closer alignment with the reference causal explanation.

Each strategy is evaluated over multiple independent trials to account for stochasticity in model responses.

\subsubsection{Results and Analysis}
\begin{figure*}
    \centering
  \includegraphics[width=5in]{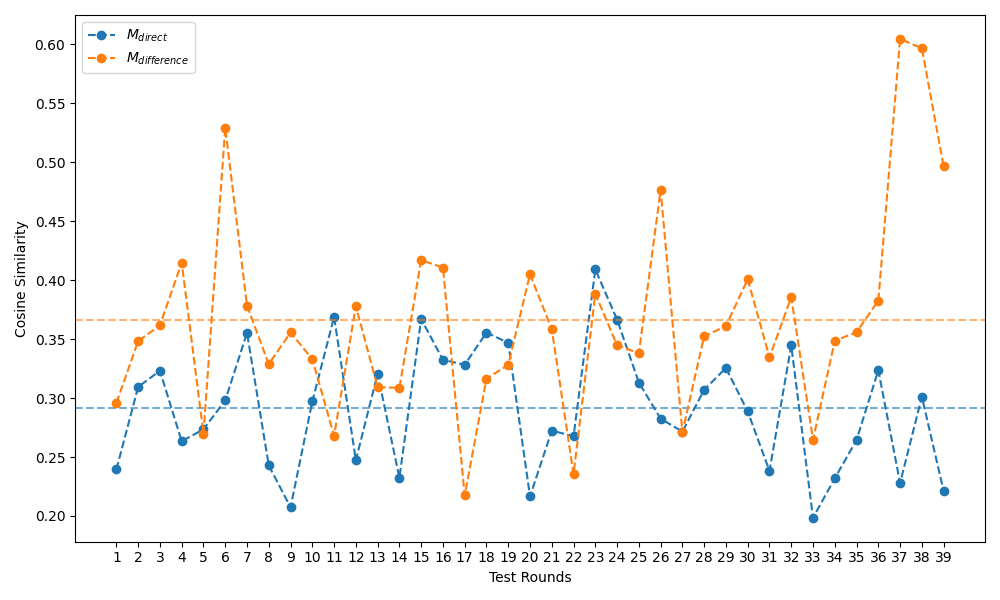}
  \caption{Comparison of semantic similarity to the reference causal explanation between the difference-oriented and direct reasoning strategies}
  \label{fig_comparison}
\end{figure*}

The experiment is conducted over 39 independent trials for both reasoning strategies using a locally deployed DeepSeek 70B model on two NVIDIA H20-3e GPUs provisioned from a GPU pool.
Figure~\ref{fig_comparison} presents the semantic similarity scores achieved by $\mathcal{M}_{\text{difference}}$ and $\mathcal{M}_{\text{direct}}$ across all trials.

$\mathcal{M}{\text{difference}}$ consistently achieves higher similarity scores than $\mathcal{M}{\text{direct}}$. The highest similarity score of 0.6044 was observed for $\mathcal{M}{\text{difference}}$ at the 37th test round. Additionally, in 15 out of 39 trials, $\mathcal{M}{\text{difference}}$ exceeded its average score of 0.3659, whereas $\mathcal{M}{\text{direct}}$ surpassed its average in only three trials (rounds 11, 23, and 24). The mean similarity scores are 0.3659 for $\mathcal{M}{\text{difference}}$ and 0.2918 for $\mathcal{M}_{\text{direct}}$, indicating a clear performance gap.

In addition to higher average similarity, the comparison-based strategy exhibits more stable performance across trials.
This suggests that explicitly reasoning in terms of differences reduces ambiguity and suppresses the influence of irrelevant contextual factors embedded in the input content.

These results demonstrate that, even in the absence of predefined measurements, difference-driven reasoning can effectively guide feedback discovery.
The experiment validates the core assumption of the proposed model: that identifying and exploiting differences provides a practical and reliable mechanism for uncovering meaningful feedback in open-ended scenarios.

\subsection{Active Action versus Passive Observation}
\label{sec_active_vs_passive}

This experiment evaluates whether \emph{active action} can accelerate causal factor identification compared with passive observation.
Unlike the previous experiment, which focuses on unknown measurement discovery, this experiment examines the efficiency of feedback acquisition under controlled environmental conditions.

\subsubsection{Simulated Environment}
\label{sec_simulated_environment}

We construct a simulated environment, denoted as $\mathcal{E}_{\text{sim}}$, consisting of a set of latent factors and observable results.
Among all factors, three factors $\{f_1, f_2, f_3\}$ are designated as \emph{effective factors}, each causally producing a unique result in a one-to-one mapping.
The remaining factors serve as \emph{disturbing factors} and do not directly generate observable results.

At each time step, the environment exposes only the currently enabled factors and their corresponding results.
The agent may query the environment state and, depending on the adopted strategy, may actively intervene by enabling or disabling selected factors.

\subsubsection{Compared Methods}
\label{sec_compared_methods}

Two agent strategies are evaluated:

\begin{itemize}
    \item \textbf{Active Method} ($\mathcal{M}_{\text{active}}$, the proposed method): the agent actively intervenes in the environment by enabling or disabling factors to induce result changes, with the goal of accelerating feedback acquisition and causal factor identification.
    \item \textbf{Observer Method} ($\mathcal{M}_{\text{observer}}$): the agent passively observes environmental changes without intervention, relying solely on randomly occurring factor changes.
\end{itemize}

In $\mathcal{M}_{\text{observer}}$, factor states change randomly at fixed time intervals, simulating a purely passive observation process.
In contrast, $\mathcal{M}_{\text{active}}$ allows the agent to select and execute interventions based on intermediate observations and reasoning outcomes.

A locally deployed \emph{DeepSeek-70B} model (the same as in \ref{sec_unknown_measurement}) is used as the large language model (LLM) for causal reasoning.
At each query, the current environment state is provided to the LLM to infer which factor causes the target result (denoted as $r_2$ in this experiment).
To avoid redundant reasoning, environment states identical to previously queried states are not resubmitted to the LLM.
In the active method, the LLM is additionally queried to suggest intervention actions (i.e., factor enabling or disabling) alongside causal judgments.

\subsubsection{Evaluation Metric}
\label{sec_eval_metric}

The primary evaluation metric is the number of LLM queries required to correctly identify the causal factor corresponding to a given target result.
A smaller number of queries indicates faster and more efficient feedback acquisition.

Each method is evaluated over multiple independent trials to account for environmental randomness and variability in LLM reasoning.

\subsubsection{Results and Analysis}
\label{sec_active_results}

\begin{figure*}
    \centering
  \includegraphics[width=5in]{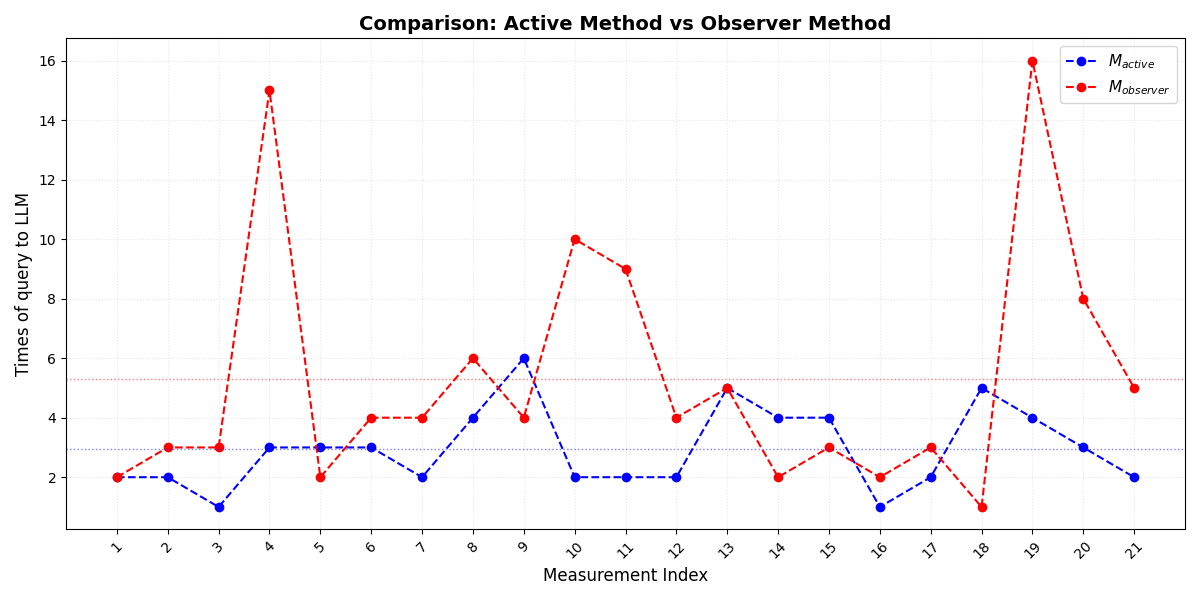}
  \caption{Comparison of query efficiency between the active method and passive observation across trials}
  \label{fig_activeMethodCmpFigure}
\end{figure*}

Figure~\ref{fig_activeMethodCmpFigure} shows the number of LLM queries required by the two methods across 20 independent trials.
The horizontal axis represents the trial index, while the vertical axis denotes the number of queries needed to correctly identify the causal factor associated with the target result.

Overall, the observer method ($\mathcal{M}_{\text{observer}}$) exhibits several large peaks in query count (e.g., 15 and 16 queries), whereas the active method ($\mathcal{M}_{\text{active}}$) maintains a substantially lower maximum (6 queries).
This behavior arises because $\mathcal{M}_{\text{observer}}$ relies on random environmental changes and must wait for informative states to appear, which can significantly delay causal identification in some trials.

In contrast, $\mathcal{M}_{\text{active}}$ actively intervenes in the environment, inducing informative changes that accelerate feedback acquisition.
As a result, its query count distribution is more compact and stable.
Quantitatively, the standard deviation of query counts for $\mathcal{M}_{\text{observer}}$ is 4.137, compared with 1.359 for $\mathcal{M}_{\text{active}}$.
The average number of queries required is 5.286 for $\mathcal{M}_{\text{observer}}$ and 2.952 for $\mathcal{M}_{\text{active}}$, further highlighting the efficiency advantage of active intervention.

\subsubsection{Statistical Significance Analysis}
To examine whether the observed difference between the two methods is statistically significant, a two-sample Welch’s $t$-test is conducted on the query counts.

The active method has a mean value of $\mu_1 = 2.95$ with a standard deviation of $\sigma_1 = 1.36$, while the observer method has a mean value of $\mu_2 = 5.29$ with a standard deviation of $\sigma_2 = 4.14$.
The test yields a $t$-statistic of $t = -2.46$ and a corresponding $p$-value of $p = 0.0216$.

Since the $p$-value is smaller than the significance level $\alpha = 0.05$, the null hypothesis of equal means is rejected.
This result indicates that the observer method requires a significantly larger number of queries than the active method, confirming the statistical significance of the performance gap.

\subsubsection{Discussion}
The above results demonstrate that enabling the agent to actively intervene in the environment substantially shortens the feedback acquisition and causal identification process.
By inducing informative environmental changes, the active method reduces reliance on random observations and mitigates delays inherent to passive strategies.

Moreover, in realistic environments with a larger number of interacting factors, passive observation is likely to become increasingly inefficient.
The results suggest that active feedback intervention is particularly important for scalable and autonomous reasoning systems, especially when combined with large language models that operate in an input-triggered manner rather than continuous perception.

\section{Conclusion} \label{sec_conclusion}

This paper addresses the limitation of predefined-measurement-based feedback mechanisms in real-world AI agents and proposes an \emph{Actively Feedback Getting} model for open-ended environments.
By treating action-induced environmental differences as the primary carrier of feedback, the proposed approach enables agents to discover, screen, and validate feedback without relying on explicit measurement definitions.
Active action is further introduced as a feedback intervention mechanism, allowing agents to deliberately manipulate environmental conditions to expose and accelerate meaningful feedback.

Through accumulated learning of action--feedback relationships, the model supports continual adaptation and reuse of prior experience.
Experimental results demonstrate that the proposed approach improves both the ability to discover unknown feedback and the efficiency of causal factor identification compared with passive observation.
These findings indicate that active feedback acquisition is a practical and effective step toward more autonomous and adaptive intelligent agents.

Future work will explore extending the model to richer real-world environments, integrating multi-agent interactions, and formalizing the theoretical properties of active feedback acquisition.


\ifCLASSOPTIONcaptionsoff
  \newpage
\fi

\bibliographystyle{IEEEtran}
\bibliography{ref}

%

\begin{IEEEbiography}{Hong Su}
  received the MS and PhD degrees, in 2006 and 2022, respectively, from Sichuan University, Chengdu, China. He is currently a researcher of Chengdu University of Information Technology Chengdu, China. His research interests include blockchain, cross-chain and smart contract.
\end{IEEEbiography}




\end{document}